\documentclass{article}
\usepackage[eabstract]{log_2026}

\usepackage{booktabs}						
\usepackage{multirow}						
\usepackage{amsfonts}						
\usepackage{graphicx}						
\usepackage[numbers,compress,sort]{natbib}	
\usepackage{enumitem}						

\newcommand{\sg}{\textsc{sg}}
\newcommand{\sa}{\textsc{sa}}
\newcommand{\ticl}{TabICL}

\newif\ifrevmarks\revmarksfalse
\DeclareRobustCommand{\rev}[1]{\ifrevmarks\textcolor{blue}{#1}\else#1\fi}
\newenvironment{revblock}{\ifrevmarks\color{blue}\fi}{}
\newcommand{\ms}[2]{#1\,{\scriptsize$\pm$#2}}   

\title[Prior-Data Fitted Relational Models via Latent Multi-table Representations]{Path to Prior-Data Fitted Relational Models via Learned Latent Multi-table Representations}

\author[M.\,S. Abolhasani et al.]{%
Mohammad Sadeq Abolhasani\\
SAP Labs\\
\email{Rastin.Abolhasani@sap.com}\And
Viswanath Ganapathy\\
SAP Labs\\
\email{viswa.ganapathy@sap.com}
}

\begin{document}

\maketitle

\begin{abstract}
Prior-data fitted networks (PFNs) show that pretraining on large populations of synthetic tasks can enable generalization to unseen datasets with little or no downstream fine-tuning. This paradigm is effective for transformer-based tabular and relational foundation models such as RDB-PFN, but remains largely unexplored for graph-based relational models built on message passing. We pretrain the graph-based relational foundation model Griffin on large collections of synthetic relational databases and ask whether curriculum learning, which benefits transformer-based relational learners, also improves message-passing representations. Rather than fine-tune on downstream tasks, we use the pretrained encoder as a general-purpose relational representation learner and pass its latent representations to a frozen in-context predictor (\ticl{}); no component is trained on the target task. On 16 real-world relational classification tasks, evaluated over five seeds against raw-feature and untrained-encoder baselines, a purely synthetic curriculum that begins with schema-guided data and continues with schema-agnostic data gives the strongest frozen in-context performance\rev{, outperforming real-data-only pretraining and every unordered single-stage mixture of the corpora; when appended after real-data pretraining, it yields the best encoders overall}. These results provide initial evidence that PFN-style synthetic pretraining transfers to graph-based relational foundation models.
\end{abstract}

\section{Introduction}
\label{sec:intro}

Relational databases (RDBs) power most enterprise workloads, yet relational foundation models still lag behind their text and vision counterparts. A central obstacle is data: high-quality RDBs are private, scarce, and structurally heterogeneous, precluding internet-scale pretraining. Prior-data fitted networks (PFNs)~\citep{muller2021transformers,hollmann2022tabpfn} address this scarcity by pretraining on an effectively unlimited population of synthetic tasks. Transformer-based models such as RDB-PFN~\citep{wang2026relational} have recently extended this idea from single tables to full relational databases. Whether the same paradigm transfers to graph-based relational foundation models built on message passing, which encode the schema's foreign-key (FK) structure as an explicit inductive bias~\citep{fey2023relational,wang2025griffin}, remains largely unexplored.

We study this question by pretraining Griffin~\citep{wang2025griffin}, a graph-centric relational foundation model, on large corpora of synthetic relational databases generated with PluRel-style pipelines~\citep{abolhasani2026plurel}. We then evaluate the resulting encoders as frozen relational representation learners that feed a frozen in-context tabular predictor, \ticl{}~\citep{qu2025tabicl}. Concretely, we ask:
\begin{enumerate}[label=\textbf{RQ\arabic*.},leftmargin=3.2em,itemsep=0.1em,topsep=0.2em]
    \item Does synthetic pretraining data help a message-passing relational encoder at all?
    \item How does synthetic pretraining compare to pretraining on real relational data?
    \item Does curriculum learning~\citep{bengio2009curriculum,abolhasani2026curriculum} over the synthetic distribution improve the learned representations?
    \item What is the role of schema-guided versus schema-agnostic synthetic data?
    \item Does more pretraining data help?
\end{enumerate}
Across seven pretraining curricula and 16 real-world relational classification tasks\rev{, extended with three unordered controls, two baselines, and five seeds per configuration,} our results answer all five questions in the affirmative, with \rev{two nuances}. Schema-guided data performs poorly on its own but provides a useful first stage before diverse schema-agnostic data. \rev{Although the purely synthetic curriculum outperforms real-data-only pretraining, the strongest encoders arise when it follows real-data pretraining, indicating that synthetic and real data are complementary.}

\section{Method}
\label{sec:method}

\paragraph{PFN-style pretraining for relational data.}
A PFN approximates Bayesian prediction by training on tasks sampled from a prior over datasets and adapting to new datasets in a single forward pass~\citep{muller2021transformers}. RDB-PFN~\citep{wang2026relational} instantiates this idea for RDBs using a transformer over inputs linearized by Deep Feature Synthesis (DFS). Each entity is flattened with a fixed bank of task-agnostic aggregations (mean, max, mode, sum, and so on) over related rows. This linearization becomes intractable as the hop count grows and, even at low hop counts, can compute many aggregations irrelevant to the downstream task. We discuss RDB-PFN not as a competing baseline, but as evidence that the PFN paradigm works for transformer-based relational models. We ask whether it also works when a message-passing architecture consumes relational structure natively and learns task-relevant multi-hop aggregation rather than enumerating it.

\paragraph{Griffin as a latent relational representation learner.}
We use Griffin~\citep{wang2025griffin}, a heterogeneous message-passing network (MPNN) over the FK graph of an RDB, as our relational encoder. Given a temporal subgraph around a target row, Griffin performs relation-aware message passing with reversed FK edges and produces a latent target-entity embedding that summarizes its multi-table context. We pretrain Griffin on supervised prediction tasks from collections of synthetic databases, giving the encoder an inductive bias toward predictive multi-table structure. Unlike RDB-PFN, which adapts a transformer over DFS features to new databases in context, our pipeline separates representation learning from prediction: Griffin learns latent multi-table representations, while a frozen in-context head predicts on unseen data.

\paragraph{Frozen in-context prediction with \ticl{}\@.}
\ticl{}~\citep{qu2025tabicl} is a prior-fitted tabular model that predicts query labels from an in-context set of labeled examples without gradient updates. We keep it fully frozen and use it as both the prediction head and evaluation engine. For each downstream task, we embed rows with the frozen pretrained Griffin encoder, provide the resulting representation--label pairs as context (sizes $\{512, 1024, 2048\}$), and let \ticl{} predict held-out labels. Because no component is trained on the target task, downstream performance directly reflects the quality of the pretrained relational representations. This yields an end-to-end prior-data-fitted pipeline for RDBs: synthetic encoder pretraining followed by frozen in-context prediction, with no real relational data required for training. \rev{Two controls using the same frozen head---raw target-row features and an untrained Griffin---anchor this measurement (Section~\ref{sec:setup}).}

\section{Experimental Setup}
\label{sec:setup}

\paragraph{Synthetic pretraining corpora.}
We generate two complementary corpora (Table~\ref{tab:corpora}). \textbf{Schema-guided (\sg{})}: 500 databases with synthetic content on a fixed real-world SAP sales (SALT) schema. \rev{The schema is the four-table FK graph of the real \texttt{rel-salt} database; PluRel generates the content, and rows are split temporally.} Each database contains 7 genuine multi-class classification tasks on real categorical sales fields: 3 at the item level and 4 at the document level, including tasks with many classes. \textbf{Schema-agnostic (\sa{})}: 9{,}143 databases from three generator sub-families, B (5{,}623), C (1{,}408), and D (2{,}112). \rev{All three use the same PluRel generator without a fixed schema and differ only in size and depth settings.} They share one task template: mask a feature column of an entity table and predict it from the row's remaining columns and its FK neighborhood. Each task is binary classification, with roughly 6 tasks per database. B, C, and D differ in structure rather than task type. B is small and shallow (2 to 4 tables, single FK depth); C has the smallest schemas, often only 2 tables; and D is large and dense (5 to 8 tables, about 12 relationships, with many FKs converging on a hub and additional cross-table FKs), placing the greatest demands on multi-hop message passing. Thus, \sg{} provides task and semantic realism on a fixed schema, whereas \sa{} provides structural diversity across schemas of increasing relational complexity. \rev{Both corpora can be regenerated exactly from integer seeds (Appendix~\ref{app:data}).}

\paragraph{Downstream tasks.}
We evaluate on 16 real-world relational classification tasks drawn from RelBench~\citep{robinson2024relbench} and 4DBInfer~\citep{wang20244dbinfer}, grouped into four families: \textbf{Com1} (diginetica-downsample-ctr, rel-hm-user-churn, retailrocket-cvr, seznam-charge, seznam-prepay), \textbf{Com2} (amazon-churn, outbrain-small-ctr, rel-avito-user-clicks, rel-avito-user-visits), \textbf{Othr1} (rel-f1-driver-dnf, rel-f1-driver-top3, stackexchange-churn, stackexchange-upvote, virus-wnv-pred), and \textbf{Othr2} (airbnb-destination, rel-trial-study-outcome). All results are ROC-AUC under frozen in-context prediction. \rev{The context is $c$ labeled training rows and the queries are the held-out rows ($c{=}1024$ here; Appendix~\ref{app:execution}).}

\paragraph{Curricula\rev{, controls, baselines,} and notation.}
We write $\mathrm{PT}_X\!\rightarrow\!\mathrm{PT}_Y$ for pretraining on corpus $X$ followed by continued pretraining on corpus $Y$, initialized from the best checkpoint of the previous stage. \rev{We write $\mathrm{PT}_{X\cup Y}$ for a single pretraining stage on the shuffled mixture of $X$ and $Y$.} $\ticl{}_c$ denotes frozen in-context evaluation with context size $c$. We compare: (i)~$\mathrm{PT}_{\sg}$, schema-guided data only; (ii)~$\mathrm{PT}_{\sa}$, schema-agnostic data only; (iii)~$\mathrm{PT}_{\sg}\!\rightarrow\!\mathrm{PT}_{\sa}$, our simple-to-diverse curriculum; (iv)~$\mathrm{PT}_{\mathrm{C1O1}}$ and $\mathrm{PT}_{\mathrm{C2O2}}$, which pretrain on real data from two task families and evaluate on the held-out two, isolating the value of real relational data; and (v)~$\mathrm{PT}_{\mathrm{C1O1}}\!\rightarrow\!\mathrm{PT}_{\sg}\!\rightarrow\!\mathrm{PT}_{\sa}$ and $\mathrm{PT}_{\mathrm{C2O2}}\!\rightarrow\!\mathrm{PT}_{\sg}\!\rightarrow\!\mathrm{PT}_{\sa}$, which append the synthetic curriculum after real-data pretraining. \rev{We add (vi)~three unordered controls, $\mathrm{PT}_{\sg\cup\sa}$, $\mathrm{PT}_{\mathrm{C1O1}\cup\sg\cup\sa}$, and $\mathrm{PT}_{\mathrm{C2O2}\cup\sg\cup\sa}$, which train once on a shuffled mixture of all 500 \sg{} databases, a fixed random subset of 1{,}400 \sa{} databases, and, for the last two, the corresponding real families; and (vii)~two baselines with the identical frozen head, \textsc{RawFeat} (\ticl{} on the raw target-row features, no encoder) and \textsc{RandGriffin} (an untrained Griffin).} Runs initialized on real data are always evaluated only on their held-out families, so there is no leakage.

\paragraph{\rev{Training, checkpoint selection, and seeds.}}
\rev{Each pretraining stage, whether initial or continued, is an independent run with the configuration in Table~\ref{tab:config}: a 100k-step ceiling, 5k-step floor, evaluation every 250 steps, and patience of 20--30 evaluations. Each stage ends at its best validation checkpoint. Thus, all configurations receive the same per-stage budget; unordered controls use one stage under the same rule, so ordered and unordered runs differ in data order rather than per-stage budget. We train and evaluate every configuration with five seeds, varying model initialization, mini-batch order, and downstream in-context sampling while fixing database-pool order and the validation split. We report the mean and standard deviation over the five end-to-end runs. Appendix~\ref{app:execution} gives the full protocol.}

\section{Results and Discussion}
\label{sec:results}

Table~\ref{tab:main} reports per-family and average ROC-AUC at context size 1024 for all curricula\rev{, the three unordered controls, and the two baselines, as mean and standard deviation over five seeds}. Appendix~\ref{app:pertask} provides full per-task results, matched-family comparisons, \rev{seed-variance analysis,} run-by-run and task-level observations, and additional training details. \rev{Unless stated otherwise, differences discussed below exceed 2.5 seed standard errors (Appendix~\ref{app:seeds}); context-size and patience ablations appear in Appendix~\ref{app:secondary}.}

\begin{table}[t]
\centering
\caption{Frozen in-context ROC-AUC by task family (context 1024)\rev{, mean $\pm$ standard deviation over five seeds}. Curricula initialized on real data are evaluated only on their held-out families. Avg is the mean over the run's evaluated tasks.}
\label{tab:main}
\scriptsize
\setlength{\tabcolsep}{3.5pt}
\renewcommand{\arraystretch}{0.92}
\begin{revblock}
\begin{tabular}{lcccccc}
\toprule
\textbf{Curriculum} & \textbf{Com1} & \textbf{Com2} & \textbf{Othr1} & \textbf{Othr2} & \textbf{Avg} & \textbf{Rank} \\
\midrule
$\mathrm{PT}_{\mathrm{C2O2}}\!\rightarrow\!\mathrm{PT}_{\sg}\!\rightarrow\!\mathrm{PT}_{\sa}$ & \ms{0.628}{0.013} & -- & \ms{0.716}{0.006} & -- & \ms{0.672}{0.004} & 1 \\
$\mathrm{PT}_{\mathrm{C1O1}}\!\rightarrow\!\mathrm{PT}_{\sg}\!\rightarrow\!\mathrm{PT}_{\sa}$ & -- & \ms{0.618}{0.030} & -- & \ms{0.730}{0.005} & \ms{0.655}{0.018} & 2 \\
$\mathrm{PT}_{\sg}\!\rightarrow\!\mathrm{PT}_{\sa}$ & \ms{0.584}{0.013} & \ms{0.613}{0.016} & \ms{0.709}{0.012} & \ms{0.729}{0.009} & \ms{0.648}{0.010} & 3 \\
$\mathrm{PT}_{\mathrm{C2O2}}$ & \ms{0.569}{0.005} & -- & \ms{0.699}{0.021} & -- & \ms{0.634}{0.012} & 4 \\
$\mathrm{PT}_{\mathrm{C1O1}}$ & -- & \ms{0.575}{0.018} & -- & \ms{0.716}{0.008} & \ms{0.622}{0.014} & 8 \\
$\mathrm{PT}_{\sa}$ & \ms{0.508}{0.029} & \ms{0.582}{0.019} & \ms{0.704}{0.008} & \ms{0.717}{0.007} & \ms{0.614}{0.010} & 9 \\
$\mathrm{PT}_{\sg}$ & \ms{0.549}{0.013} & \ms{0.557}{0.016} & \ms{0.644}{0.012} & \ms{0.740}{0.009} & \ms{0.605}{0.010} & 10 \\
\midrule
$\mathrm{PT}_{\mathrm{C1O1}\cup\sg\cup\sa}$ (unordered) & -- & \ms{0.598}{0.016} & -- & \ms{0.698}{0.009} & \ms{0.631}{0.010} & 5 \\
$\mathrm{PT}_{\mathrm{C2O2}\cup\sg\cup\sa}$ (unordered) & \ms{0.579}{0.013} & -- & \ms{0.677}{0.012} & -- & \ms{0.628}{0.010} & 6 \\
$\mathrm{PT}_{\sg\cup\sa}$ (unordered) & \ms{0.571}{0.013} & \ms{0.599}{0.016} & \ms{0.682}{0.012} & \ms{0.676}{0.009} & \ms{0.626}{0.010} & 7 \\
\midrule
\textsc{RandGriffin} (untrained encoder) & \ms{0.545}{0.004} & \ms{0.530}{0.007} & \ms{0.656}{0.018} & \ms{0.654}{0.013} & \ms{0.590}{0.003} & 11 \\
\textsc{RawFeat} (no encoder) & \ms{0.407}{0.020} & \ms{0.501}{0.004} & \ms{0.618}{0.014} & \ms{0.752}{0.008} & \ms{0.539}{0.008} & 12 \\
\bottomrule
\end{tabular}
\end{revblock}
\end{table}

\paragraph{Synthetic data helps (RQ1), and more data helps (RQ5).}
\rev{The two baselines fix the floor: \ticl{} on the raw target-row features reaches 0.539, an untrained Griffin 0.590, and every pretrained encoder lies above both, from 0.605 for $\mathrm{PT}_{\sg}$ to 0.672. The best purely synthetic curriculum improves on the untrained encoder by 0.058, so the gain comes from pretraining, not from the architecture or the head.} Appending the synthetic curriculum after real-data pretraining improves held-out performance in both directions of the split. $\mathrm{PT}_{\mathrm{C1O1}}\!\rightarrow\!\mathrm{PT}_{\sg}\!\rightarrow\!\mathrm{PT}_{\sa}$ outperforms $\mathrm{PT}_{\mathrm{C1O1}}$ (\rev{0.655 vs.\ 0.622} average, and \rev{0.618 vs.\ 0.575 on Com2}), and $\mathrm{PT}_{\mathrm{C2O2}}\!\rightarrow\!\mathrm{PT}_{\sg}\!\rightarrow\!\mathrm{PT}_{\sa}$ outperforms $\mathrm{PT}_{\mathrm{C2O2}}$ by a \rev{similar} margin (\rev{0.672 vs.\ 0.634} average, and \rev{0.628 vs.\ 0.569} on Com1). Since the added stages are entirely synthetic, this shows two things at once. Additional pretraining data helps, and synthetic data improves even a model that has already seen real relational data from the same downstream domain.

\paragraph{Synthetic \rev{data} outperforms real \rev{data alone} (RQ2).}
The heart of our study is $\mathrm{PT}_{\sg}\!\rightarrow\!\mathrm{PT}_{\sa}$. Pretrained on no real relational data at all, it achieves the best \rev{purely synthetic} average (\rev{0.648}) and also outperforms \rev{both runs that pretrain on real data alone} when compared on the matched held-out families (\rev{0.646 vs.\ 0.634 for $\mathrm{PT}_{\mathrm{C2O2}}$ on Com1$+$Othr1, 1.7 standard errors, and 0.652 vs.\ 0.622 for $\mathrm{PT}_{\mathrm{C1O1}}$ on Com2$+$Othr2}). Pretraining on a narrow set of real databases appears to lock the encoder onto those specific schemas and distributions\rev{. Real data is not harmful in itself, however: appending the synthetic stages after real-data pretraining gives the strongest encoders in the study. $\mathrm{PT}_{\mathrm{C2O2}}\!\rightarrow\!\mathrm{PT}_{\sg}\!\rightarrow\!\mathrm{PT}_{\sa}$ (0.672) exceeds $\mathrm{PT}_{\sg}\!\rightarrow\!\mathrm{PT}_{\sa}$ on the matched subset by 0.026, and $\mathrm{PT}_{\mathrm{C1O1}}\!\rightarrow\!\mathrm{PT}_{\sg}\!\rightarrow\!\mathrm{PT}_{\sa}$ (0.655) ties it on Com2$+$Othr2 (0.652); this is the only ordering whose direction changes once results are averaged over five seeds}.

\paragraph{Curriculum and schema guidance matter (RQ3, RQ4).}
$\mathrm{PT}_{\sg}$ alone ranks last \rev{among the pretrained encoders (0.605), only 0.015 above the untrained one}. Five hundred databases on one fixed schema yield a narrow encoder. Yet schema-guided data is far from useless. The only difference between $\mathrm{PT}_{\sa}$ (\rev{0.614}) and $\mathrm{PT}_{\sg}\!\rightarrow\!\mathrm{PT}_{\sa}$ (\rev{0.648}) is the schema-guided opening stage, and that stage delivers a consistent gain across \rev{all four families ($+$0.076 on Com1, $+$0.031 on Com2, $+$0.005 on Othr1, $+$0.012 on Othr2), clear on the commerce families and within seed noise on the Othr families}. \rev{The unordered controls show that the order of the stages, not merely the presence of both corpora, is what matters. One shuffled stage over all \sg{} and 1{,}400 \sa{} databases gives 0.626, slightly above $\mathrm{PT}_{\sa}$ on all 9{,}143 \sa{} databases (0.614) but 0.022 below the ordered curriculum and behind it on every family; after real-data initialization the unordered mixtures reach 0.631 and 0.628 against 0.655 and 0.672 for the ordered curricula, the latter no better than real data alone (0.634). Since every stage runs under the same budget and stopping rule, extra optimization cannot explain the gain (Appendix~\ref{app:runbyrun} discusses the remaining data-volume caveat).} Opening on semantically realistic tasks over a fixed real-world schema, and then broadening to structurally diverse schema-agnostic data, is an effective simple-to-diverse curriculum for message-passing encoders. This mirrors curriculum effects previously observed in transformer-based relational PFNs~\citep{abolhasani2026curriculum}.

\section{Limitations and Future Work}
\label{sec:limits}

Our study covers only classification; extending the pipeline and the \sa{} task template to regression is a natural next step. We also explore only a small curriculum space. Promising variants include placing real data after the synthetic stages ($\sg\!\rightarrow\!\sa\!\rightarrow\!$ real) and ordering \sa{} sub-families by structural complexity ($\sg\!\rightarrow\!\mathrm{B}\!\rightarrow\!\mathrm{C}\!\rightarrow\!\mathrm{D}$). \rev{Because the unordered controls use 1{,}400 rather than all 9{,}143 \sa{} databases, a full-union single-stage run and a second \sa{} stage on top of $\mathrm{PT}_{\sa}$ would complete the compute-matched comparison; per-task and supervised baselines also remain future work.} Finally, we intentionally keep the in-context head frozen. Jointly training the encoder and head end to end, as transformer-based RDB-PFN effectively does, may reduce the remaining gap between representation quality and predictor capacity. \rev{Like other relational foundation models, ours uses orders of magnitude less compute than language models; our claim is limited to transfer of the recipe at this scale.}

\bibliographystyle{unsrtnat}
\bibliography{references}

\appendix

\section{Full Per-task Results}
\label{app:pertask}

Table~\ref{tab:pertask} lists the ROC-AUC of every curriculum on every evaluated task at context size 1024\rev{, averaged over five seeds, together with the three unordered controls and the two baselines}. Empty cells mean the task family was part of that run's pretraining data and is therefore excluded from its evaluation.

\begin{table}[!htb]
\centering
\caption{Per-task ROC-AUC under frozen in-context prediction (context 1024)\rev{, mean over five seeds}. Empty cells: the family was used for pretraining and is excluded from evaluation. \rev{Columns: (1)~$\mathrm{PT}_{\mathrm{C2O2}\rightarrow\sg\rightarrow\sa}$, (2)~$\mathrm{PT}_{\mathrm{C1O1}\rightarrow\sg\rightarrow\sa}$, (3)~$\mathrm{PT}_{\sg\rightarrow\sa}$, (4)~$\mathrm{PT}_{\mathrm{C2O2}}$, (5)~$\mathrm{PT}_{\mathrm{C1O1}\cup\sg\cup\sa}$, (6)~$\mathrm{PT}_{\mathrm{C2O2}\cup\sg\cup\sa}$, (7)~$\mathrm{PT}_{\sg\cup\sa}$, (8)~$\mathrm{PT}_{\mathrm{C1O1}}$, (9)~$\mathrm{PT}_{\sa}$, (10)~$\mathrm{PT}_{\sg}$, (11)~\textsc{RandGriffin}, (12)~\textsc{RawFeat}.}}
\label{tab:pertask}
\scriptsize
\setlength{\tabcolsep}{2.6pt}
\begin{revblock}
\begin{tabular}{lcccccccccccc}
\toprule
\textbf{Task} & \textbf{(1)} & \textbf{(2)} & \textbf{(3)} & \textbf{(4)} & \textbf{(5)} & \textbf{(6)} & \textbf{(7)} & \textbf{(8)} & \textbf{(9)} & \textbf{(10)} & \textbf{(11)} & \textbf{(12)} \\
\midrule
Com1: diginetica-downsample-ctr & 0.665 & -- & 0.605 & 0.607 & -- & 0.550 & 0.596 & -- & 0.512 & 0.552 & 0.542 & 0.490 \\
Com1: rel-hm-user-churn & 0.565 & -- & 0.566 & 0.578 & -- & 0.587 & 0.565 & -- & 0.541 & 0.565 & 0.495 & 0.542 \\
Com1: retailrocket-cvr & 0.877 & -- & 0.901 & 0.858 & -- & 0.945 & 0.906 & -- & 0.773 & 0.838 & 0.835 & 0.495 \\
Com1: seznam-charge & 0.435 & -- & 0.330 & 0.270 & -- & 0.279 & 0.278 & -- & 0.252 & 0.245 & 0.351 & 0.226 \\
Com1: seznam-prepay & 0.597 & -- & 0.518 & 0.532 & -- & 0.534 & 0.510 & -- & 0.462 & 0.544 & 0.503 & 0.282 \\
\midrule
Com2: amazon-churn & -- & 0.640 & 0.623 & -- & 0.639 & -- & 0.674 & 0.613 & 0.629 & 0.657 & 0.565 & 0.500 \\
Com2: outbrain-small-ctr & -- & 0.640 & 0.607 & -- & 0.582 & -- & 0.526 & 0.574 & 0.528 & 0.517 & 0.464 & 0.501 \\
Com2: rel-avito-user-clicks & -- & 0.584 & 0.619 & -- & 0.592 & -- & 0.593 & 0.524 & 0.605 & 0.462 & 0.543 & 0.502 \\
Com2: rel-avito-user-visits & -- & 0.608 & 0.603 & -- & 0.578 & -- & 0.604 & 0.589 & 0.566 & 0.592 & 0.548 & 0.501 \\
\midrule
Othr1: rel-f1-driver-dnf & 0.676 & -- & 0.680 & 0.665 & -- & 0.664 & 0.675 & -- & 0.674 & 0.638 & 0.623 & 0.629 \\
Othr1: rel-f1-driver-top3 & 0.740 & -- & 0.772 & 0.685 & -- & 0.734 & 0.739 & -- & 0.698 & 0.586 & 0.653 & 0.605 \\
Othr1: stackexchange-churn & 0.701 & -- & 0.706 & 0.682 & -- & 0.621 & 0.705 & -- & 0.715 & 0.667 & 0.655 & 0.641 \\
Othr1: stackexchange-upvote & 0.830 & -- & 0.841 & 0.825 & -- & 0.806 & 0.828 & -- & 0.787 & 0.735 & 0.773 & 0.562 \\
Othr1: virus-wnv-pred & 0.634 & -- & 0.545 & 0.639 & -- & 0.560 & 0.463 & -- & 0.645 & 0.594 & 0.576 & 0.653 \\
\midrule
Othr2: airbnb-destination & -- & 0.853 & 0.852 & -- & 0.840 & -- & 0.842 & 0.854 & 0.845 & 0.851 & 0.767 & 0.838 \\
Othr2: rel-trial-study-outcome & -- & 0.607 & 0.606 & -- & 0.556 & -- & 0.510 & 0.578 & 0.589 & 0.629 & 0.541 & 0.666 \\
\midrule
\textbf{Average} & \textbf{0.672} & 0.655 & 0.648 & 0.634 & 0.631 & 0.628 & 0.626 & 0.622 & 0.614 & 0.605 & 0.590 & 0.539 \\
\bottomrule
\end{tabular}
\end{revblock}
\end{table}

\section{Matched-family Comparisons}
\label{app:matched}

Table~\ref{tab:main} averages each run over its own evaluated tasks, so runs initialized on real data cover fewer tasks than fully synthetic runs. For an exact comparison, Table~\ref{tab:matched} restricts all runs to the same subsets: the 10 Com1$+$Othr1 tasks and the 6 Com2$+$Othr2 tasks. On both subsets, $\mathrm{PT}_{\sg}\!\rightarrow\!\mathrm{PT}_{\sa}$ remains the best \rev{purely synthetic} curriculum. It outperforms both real-data-only runs \rev{(by 0.012 and 0.030), all three unordered controls, and both baselines. The exceptions are continuations from real-data checkpoints: $\mathrm{PT}_{\mathrm{C2O2}}\!\rightarrow\!\mathrm{PT}_{\sg}\!\rightarrow\!\mathrm{PT}_{\sa}$ leads it by 0.026 on Com1$+$Othr1, while $\mathrm{PT}_{\mathrm{C1O1}}\!\rightarrow\!\mathrm{PT}_{\sg}\!\rightarrow\!\mathrm{PT}_{\sa}$ is effectively tied on Com2$+$Othr2 (0.655 vs.\ 0.652, well within one standard deviation)}. Thus, the \rev{curriculum ordering} in Table~\ref{tab:main} is not an artifact of averaging over different task sets.

\begin{table}[!htb]
\centering
\caption{ROC-AUC averaged over fixed task subsets (context 1024)\rev{, mean $\pm$ standard deviation over five seeds}. Each column compares all runs that were evaluated on that subset.}
\label{tab:matched}
\small
\begin{revblock}
\begin{tabular}{lcc}
\toprule
\textbf{Curriculum} & \textbf{Com1$+$Othr1 (10 tasks)} & \textbf{Com2$+$Othr2 (6 tasks)} \\
\midrule
$\mathrm{PT}_{\mathrm{C2O2}}\!\rightarrow\!\mathrm{PT}_{\sg}\!\rightarrow\!\mathrm{PT}_{\sa}$ & \textbf{\ms{0.672}{0.004}} & -- \\
$\mathrm{PT}_{\mathrm{C1O1}}\!\rightarrow\!\mathrm{PT}_{\sg}\!\rightarrow\!\mathrm{PT}_{\sa}$ & -- & \textbf{\ms{0.655}{0.018}} \\
$\mathrm{PT}_{\sg}\!\rightarrow\!\mathrm{PT}_{\sa}$ & \ms{0.646}{0.010} & \ms{0.652}{0.010} \\
$\mathrm{PT}_{\mathrm{C2O2}}$ & \ms{0.634}{0.012} & -- \\
$\mathrm{PT}_{\mathrm{C1O1}}$ & -- & \ms{0.622}{0.014} \\
$\mathrm{PT}_{\sa}$ & \ms{0.606}{0.011} & \ms{0.627}{0.013} \\
$\mathrm{PT}_{\sg}$ & \ms{0.596}{0.010} & \ms{0.618}{0.010} \\
\midrule
$\mathrm{PT}_{\mathrm{C1O1}\cup\sg\cup\sa}$ (unordered) & -- & \ms{0.631}{0.010} \\
$\mathrm{PT}_{\mathrm{C2O2}\cup\sg\cup\sa}$ (unordered) & \ms{0.628}{0.010} & -- \\
$\mathrm{PT}_{\sg\cup\sa}$ (unordered) & \ms{0.627}{0.010} & \ms{0.625}{0.010} \\
\midrule
\textsc{RandGriffin} (untrained encoder) & \ms{0.601}{0.009} & \ms{0.571}{0.007} \\
\textsc{RawFeat} (no encoder) & \ms{0.512}{0.014} & \ms{0.585}{0.003} \\
\bottomrule
\end{tabular}
\end{revblock}
\end{table}

\section[Seed Variance and Reliability of the Comparisons]{\rev{Seed Variance and Reliability of the Comparisons}}
\label{app:seeds}

\begin{revblock}
Every configuration was trained and evaluated five times with different seeds, and Tables~\ref{tab:main} and~\ref{tab:matched} report the standard deviation of the task-averaged ROC-AUC over these five end-to-end runs. Seed standard deviations lie between 0.003 and 0.018 for the averages, and between 0.004 and 0.030 for individual families; Com2 is the noisiest family. Table~\ref{tab:seeds} lists the comparisons that carry our conclusions, with the difference of means $\Delta$, the standard error of the difference $\mathrm{SE}=\sqrt{(s_1^2+s_2^2)/5}$, and their ratio. As a rough guide, a ratio above 2.3 corresponds to $p<0.05$ in a two-sided $t$ test with eight degrees of freedom, and a ratio above 2.8 to the same level under the most conservative Welch degrees of freedom. All comparisons on which Section~\ref{sec:results} relies exceed 2.5 except two, which the text reports as such: the purely synthetic curriculum against $\mathrm{PT}_{\mathrm{C2O2}}$ on the matched subset (ratio 1.7) and the comparison between $\mathrm{PT}_{\mathrm{C1O1}}\!\rightarrow\!\mathrm{PT}_{\sg}\!\rightarrow\!\mathrm{PT}_{\sa}$ and $\mathrm{PT}_{\sg}\!\rightarrow\!\mathrm{PT}_{\sa}$ on Com2$+$Othr2 (ratio 0.3, a tie). The unordered mixture of real and synthetic data in the C2O2 direction is likewise indistinguishable from real data alone (ratio $-0.9$), the advantage of the unordered synthetic mixture over $\mathrm{PT}_{\sa}$ is suggestive rather than conclusive (ratio 1.9), and differences of less than about 0.015 between individual entries of Table~\ref{tab:main}, such as $\mathrm{PT}_{\sg}$ against $\mathrm{PT}_{\sa}$ or the Othr1 and Othr2 gains of the opening stage, are within seed noise and are not interpreted on their own.
\end{revblock}

\begin{table}[!htb]
\centering
\caption{\rev{Key comparisons with their seed standard error. $\Delta$ is the difference of the two means on the task set in the last column; SE is the standard error of the difference over five seeds per arm.}}
\label{tab:seeds}
\scriptsize
\setlength{\tabcolsep}{4pt}
\begin{revblock}
\begin{tabular}{llcccl}
\toprule
\textbf{Conclusion} & \textbf{Comparison} & $\boldsymbol{\Delta}$ & \textbf{SE} & $\boldsymbol{\Delta/\mathrm{SE}}$ & \textbf{Tasks} \\
\midrule
Pretraining helps (RQ1) & $\mathrm{PT}_{\sg\rightarrow\sa}$ vs.\ \textsc{RandGriffin} & $+$0.058 & 0.005 & 12.4 & all 16 \\
 & $\mathrm{PT}_{\sg}$ vs.\ \textsc{RandGriffin} & $+$0.015 & 0.005 & 3.2 & all 16 \\
 & \textsc{RandGriffin} vs.\ \textsc{RawFeat} & $+$0.051 & 0.004 & 13.3 & all 16 \\
More data helps (RQ5) & $\mathrm{PT}_{\mathrm{C1O1}\rightarrow\sg\rightarrow\sa}$ vs.\ $\mathrm{PT}_{\mathrm{C1O1}}$ & $+$0.033 & 0.010 & 3.2 & Com2$+$Othr2 \\
 & $\mathrm{PT}_{\mathrm{C2O2}\rightarrow\sg\rightarrow\sa}$ vs.\ $\mathrm{PT}_{\mathrm{C2O2}}$ & $+$0.038 & 0.006 & 6.7 & Com1$+$Othr1 \\
Synthetic vs.\ real alone (RQ2) & $\mathrm{PT}_{\sg\rightarrow\sa}$ vs.\ $\mathrm{PT}_{\mathrm{C2O2}}$ & $+$0.012 & 0.007 & 1.7 & Com1$+$Othr1 \\
 & $\mathrm{PT}_{\sg\rightarrow\sa}$ vs.\ $\mathrm{PT}_{\mathrm{C1O1}}$ & $+$0.030 & 0.008 & 3.9 & Com2$+$Othr2 \\
Real then synthetic (RQ2) & $\mathrm{PT}_{\mathrm{C2O2}\rightarrow\sg\rightarrow\sa}$ vs.\ $\mathrm{PT}_{\sg\rightarrow\sa}$ & $+$0.026 & 0.005 & 5.4 & Com1$+$Othr1 \\
 & $\mathrm{PT}_{\mathrm{C1O1}\rightarrow\sg\rightarrow\sa}$ vs.\ $\mathrm{PT}_{\sg\rightarrow\sa}$ & $+$0.003 & 0.009 & 0.3 & Com2$+$Othr2 \\
Opening stage (RQ3, RQ4) & $\mathrm{PT}_{\sg\rightarrow\sa}$ vs.\ $\mathrm{PT}_{\sa}$ & $+$0.034 & 0.006 & 5.4 & all 16 \\
Stage order (RQ3) & $\mathrm{PT}_{\sg\rightarrow\sa}$ vs.\ $\mathrm{PT}_{\sg\cup\sa}$ & $+$0.022 & 0.006 & 3.5 & all 16 \\
 & $\mathrm{PT}_{\sg\cup\sa}$ vs.\ $\mathrm{PT}_{\sa}$ & $+$0.012 & 0.006 & 1.9 & all 16 \\
 & $\mathrm{PT}_{\mathrm{C1O1}\rightarrow\sg\rightarrow\sa}$ vs.\ $\mathrm{PT}_{\mathrm{C1O1}\cup\sg\cup\sa}$ & $+$0.024 & 0.009 & 2.6 & Com2$+$Othr2 \\
 & $\mathrm{PT}_{\mathrm{C2O2}\rightarrow\sg\rightarrow\sa}$ vs.\ $\mathrm{PT}_{\mathrm{C2O2}\cup\sg\cup\sa}$ & $+$0.044 & 0.005 & 9.1 & Com1$+$Othr1 \\
 & $\mathrm{PT}_{\mathrm{C2O2}\cup\sg\cup\sa}$ vs.\ $\mathrm{PT}_{\mathrm{C2O2}}$ & $-$0.006 & 0.007 & $-$0.9 & Com1$+$Othr1 \\
\bottomrule
\end{tabular}
\end{revblock}
\end{table}

\section{Run-by-run Observations}
\label{app:runbyrun}

\paragraph{\rev{$\mathrm{PT}_{\mathrm{C2O2}}$ (rank 4, avg 0.634) and $\mathrm{PT}_{\mathrm{C2O2}}\!\rightarrow\!\mathrm{PT}_{\sg}\!\rightarrow\!\mathrm{PT}_{\sa}$ (rank 1, avg 0.672).}}
The mirrored split behaves consistently, with a larger margin. Appending the synthetic curriculum improves the average from \rev{0.634 to 0.672} and lifts Com1 from \rev{0.569 to 0.628}, including sizable per-task gains on diginetica-downsample-ctr (\rev{$+$0.058}), seznam-charge (\rev{$+$0.165}), and seznam-prepay (\rev{$+$0.065})\rev{, and Othr1 from 0.699 to 0.716, led by rel-f1-driver-top3 ($+$0.055)}. As in the C1O1 direction, the improvement is not uniform: the continuation dips \rev{slightly on rel-hm-user-churn ($-$0.013)} and \rev{on virus-wnv-pred ($-$0.005)}. \rev{The continuation is the strongest run in the study and the most stable across seeds ($\pm$0.004). On the matched Com1$+$Othr1 subset it exceeds $\mathrm{PT}_{\sg}\!\rightarrow\!\mathrm{PT}_{\sa}$ by 0.026 (0.672 vs.\ 0.646), whereas real data alone falls 0.012 short of the purely synthetic curriculum (0.634 vs.\ 0.646).}

\paragraph{$\mathrm{PT}_{\mathrm{C1O1}}$ (rank \rev{8}, avg \rev{0.622}) and $\mathrm{PT}_{\mathrm{C1O1}}\!\rightarrow\!\mathrm{PT}_{\sg}\!\rightarrow\!\mathrm{PT}_{\sa}$ (rank \rev{2}, avg \rev{0.655}).}
The real-data-only run establishes what pretraining on Com1$+$Othr1 alone transfers to the held-out Com2$+$Othr2 families. Appending the synthetic curriculum improves the average from \rev{0.622 to 0.655}, driven \rev{by Com2 (0.575 to 0.618) as much as by Othr2 (0.716 to 0.730), with gains on every task except airbnb-destination (unchanged at 0.85): outbrain-small-ctr ($+$0.066), rel-avito-user-clicks ($+$0.060), rel-trial-study-outcome ($+$0.029), amazon-churn ($+$0.027), and rel-avito-user-visits ($+$0.019)}. This supports two conclusions at once: more pretraining data helps, and synthetic data helps even for a model that has already seen real relational data. \rev{On the matched Com2$+$Othr2 subset the continuation ($0.655\pm0.018$) and the purely synthetic $\mathrm{PT}_{\sg}\!\rightarrow\!\mathrm{PT}_{\sa}$ ($0.652\pm0.010$) are statistically tied. This is also the run with the largest seed variation, concentrated on Com2 ($\pm$0.030).}

\paragraph{$\mathrm{PT}_{\sg}\!\rightarrow\!\mathrm{PT}_{\sa}$ (rank \rev{3}, avg \rev{0.648}).}
This run is the heart of our experiments. With no real relational data in pretraining, it achieves the best average of \rev{the four fully synthetic configurations and of all runs evaluated on all 16 tasks}, and it stays \rev{ahead of every real-data-only run, every unordered control, and both baselines} when restricted to either matched task subset (Table~\ref{tab:matched}). It is the strongest \rev{fully synthetic run on Com1 (0.584), Com2 (0.613), and Othr1 (0.709), and within 0.011 of $\mathrm{PT}_{\sg}$ on Othr2 (0.729 vs.\ 0.740)}. Among the \rev{four} fully synthetic runs it wins the most individual tasks: \rev{9} of 16, against \rev{3 for the unordered $\mathrm{PT}_{\sg\cup\sa}$, 2} for $\mathrm{PT}_{\sa}$ and \rev{2} for $\mathrm{PT}_{\sg}$. Its largest single-task leads over $\mathrm{PT}_{\sa}$ are retailrocket-cvr (\rev{$+$0.128}), \rev{diginetica-downsample-ctr ($+$0.093), outbrain-small-ctr ($+$0.079), seznam-charge ($+$0.078), and rel-f1-driver-top3 ($+$0.074); its one sizable loss is virus-wnv-pred ($-$0.100)}.

\paragraph{\rev{Unordered controls: $\mathrm{PT}_{\mathrm{C1O1}\cup\sg\cup\sa}$ (rank 5, avg 0.631), $\mathrm{PT}_{\mathrm{C2O2}\cup\sg\cup\sa}$ (rank 6, avg 0.628), and $\mathrm{PT}_{\sg\cup\sa}$ (rank 7, avg 0.626).}}
\rev{Each unordered control trails its ordered counterpart on every family: by 0.022 on average for the purely synthetic pair (0.626 vs.\ 0.648; Com1 0.571 vs.\ 0.584, Com2 0.599 vs.\ 0.613, Othr1 0.682 vs.\ 0.709, Othr2 0.676 vs.\ 0.729), by 0.024 for the C1O1 pair (0.631 vs.\ 0.655), and by 0.044 for the C2O2 pair (0.628 vs.\ 0.672), where mixing real and synthetic data in one stage is no better than real data alone (0.634). Mixing does help relative to the single corpora: $\mathrm{PT}_{\sg\cup\sa}$ is above both $\mathrm{PT}_{\sa}$ (0.614) and $\mathrm{PT}_{\sg}$ (0.605) although it sees only 1{,}400 of the 9{,}143 \sa{} databases, so the volume of schema-agnostic data alone does not explain the gap to the ordered curriculum; a single-stage run on the full union would settle the remaining confound and is left for future work. The unordered mixture wins three individual tasks among the fully synthetic runs (retailrocket-cvr 0.906, amazon-churn 0.674, rel-avito-user-visits 0.604) but is the weakest run of all on virus-wnv-pred (0.463) and rel-trial-study-outcome (0.510).}

\paragraph{$\mathrm{PT}_{\sa}$ (rank \rev{9}, avg \rev{0.614}).}
Schema-agnostic data alone \rev{produces a clearly better encoder than the untrained one (0.614 vs.\ 0.590) and is the best fully synthetic run on stackexchange-churn (0.715) and virus-wnv-pred (0.645), but it no longer exceeds the runs initialized on real data: with five seeds it falls below $\mathrm{PT}_{\mathrm{C2O2}}$ on Com1$+$Othr1 (0.606 vs.\ 0.634) and is level with $\mathrm{PT}_{\mathrm{C1O1}}$ on Com2$+$Othr2 (0.627 vs.\ 0.622). It is the weakest of all pretrained runs on Com1 (0.508), where seznam-charge (0.252) and seznam-prepay (0.462) pull it down}. The gap to $\mathrm{PT}_{\sg}\!\rightarrow\!\mathrm{PT}_{\sa}$ isolates the value of the schema-guided stage, since the two runs are otherwise identical: the opening stage adds \rev{$+$0.076} on Com1, \rev{$+$0.031} on Com2, \rev{$+$0.005} on Othr1, \rev{and $+$0.012 on Othr2}.

\paragraph{$\mathrm{PT}_{\sg}$ (rank \rev{10}, avg \rev{0.605}).}
Pretraining only on the 500 schema-guided SALT databases yields the weakest \rev{pretrained} encoder\rev{, only 0.015 above the untrained baseline}. All the data lives on one fixed schema, so the encoder develops a narrow understanding of that specific structure and transfers poorly, with the weakest results of any \rev{pretrained} run on Com2 (\rev{0.557}) and Othr1 (\rev{0.644}). This does not make schema-guided data useless. Used as the opening stage before the schema-agnostic corpus, it improves \rev{all four families} (see the $\mathrm{PT}_{\sa}$ paragraph above). The correct reading is that schema-guided data is a curriculum ingredient, not a standalone corpus. Interestingly, $\mathrm{PT}_{\sg}$ is competitive or best on a few individual tasks (amazon-churn at \rev{0.657}, rel-trial-study-outcome at \rev{0.629}, airbnb-destination at \rev{0.851})\rev{, and it has the best Othr2 average of any pretrained run (0.740)}, which suggests those tasks reward the semantic realism of the SALT tasks more than structural diversity.

\paragraph{\rev{Baselines: \textsc{RandGriffin} (rank 11, avg 0.590) and \textsc{RawFeat} (rank 12, avg 0.539).}}
\rev{An untrained Griffin already improves on the raw target-row features by 0.051 on average and by 0.138 on Com1 (0.545 vs.\ 0.407), which shows that even random message passing carries neighborhood information that the row itself lacks; it is also remarkably stable across initializations ($\pm$0.003). The raw-feature baseline is at chance on all four Com2 tasks (0.500 to 0.502) and on retailrocket-cvr (0.495), so on the commerce tasks the discriminative signal is relational and can only enter through the encoder. The one family where raw features are competitive is Othr2 (0.752, the highest of any run), driven by rel-trial-study-outcome (0.666), whose label appears to be largely predictable from the row itself. Every pretrained encoder exceeds the untrained one on its own task set, by 0.015 ($\mathrm{PT}_{\sg}$) up to 0.084 ($\mathrm{PT}_{\mathrm{C1O1}}\!\rightarrow\!\mathrm{PT}_{\sg}\!\rightarrow\!\mathrm{PT}_{\sa}$ on Com2$+$Othr2, where the untrained encoder scores 0.571).}

\section{Task-level Observations}
\label{app:tasklevel}

\paragraph{seznam-charge is an outlier for every curriculum.}
All \rev{nine} runs evaluated on Com1 score below 0.5 on seznam-charge (\rev{0.226 to 0.435}). Since a consistent below-chance AUROC indicates a systematically inverted ranking rather than an uninformative one, this points to a mismatch between the in-context examples and the query rows for this task, for example a temporal distribution shift, rather than to a failure of any particular pretraining recipe. \rev{That the raw-feature baseline is the lowest of all (0.226) and the untrained encoder is also inverted (0.351) confirms that the inversion originates in the task's rows and split rather than in any encoder.} The task depresses the Com1 average of every run equally, so it does not change any ranking, but the family averages should be read with it in mind. Notably, the \rev{two strongest curricula} are the least affected ($\mathrm{PT}_{\mathrm{C2O2}}\!\rightarrow\!\mathrm{PT}_{\sg}\!\rightarrow\!\mathrm{PT}_{\sa}$ at \rev{0.435} and $\mathrm{PT}_{\sg\rightarrow\sa}$ at \rev{0.330}, versus \rev{0.245} for $\mathrm{PT}_{\sg}$ \rev{and 0.252 for $\mathrm{PT}_{\sa}$}).

\paragraph{airbnb-destination is insensitive to pretraining.}
Every \rev{pretrained} run scores between \rev{0.840 and 0.854} on airbnb-destination\rev{, and so does the raw-feature baseline (0.838); only the untrained encoder falls to 0.767}. The signal for this task appears to be captured by the frozen \ticl{} head from almost any reasonable representation, so it contributes little to separating the curricula.

\paragraph{Where the curricula separate.}
The tasks that discriminate most strongly between \rev{pretrained} curricula are retailrocket-cvr (spread \rev{0.773 to 0.945} across runs), rel-f1-driver-top3 (\rev{0.586 to 0.772}), \rev{rel-avito-user-clicks (0.462 to 0.619), diginetica-downsample-ctr (0.512 to 0.665), and virus-wnv-pred (0.463 to 0.645), in addition to the seznam-charge outlier above}. \rev{Most} of these are tasks where multi-hop relational context is plausibly informative, which is consistent with the encoder, rather than the frozen head, being the component that the curricula differentiate. \rev{The raw-feature baseline is at chance on two of them (retailrocket-cvr 0.495, rel-avito-user-clicks 0.502) and far below every pretrained run on stackexchange-upvote (0.562 vs.\ 0.735 to 0.841), which confirms that the discriminating signal enters through the relational encoder.}

\paragraph{\rev{virus-wnv-pred and rel-trial-study-outcome favor less synthetic pretraining.}}
\rev{On virus-wnv-pred the best pretrained runs are $\mathrm{PT}_{\sa}$ (0.645), $\mathrm{PT}_{\mathrm{C2O2}}$ (0.639), and $\mathrm{PT}_{\mathrm{C2O2}}\!\rightarrow\!\mathrm{PT}_{\sg}\!\rightarrow\!\mathrm{PT}_{\sa}$ (0.634), while the schema-guided opening stage costs 0.100 ($\mathrm{PT}_{\sg\rightarrow\sa}$ at 0.545) and the unordered mixture 0.182 (0.463). On rel-trial-study-outcome the raw features are best (0.666) and every encoder loses some of the row's own signal. These two tasks are the main sources of per-task disagreement with the average ranking, and both belong to the Othr families, where relational context appears to matter least.}

\section{Secondary Ablations}
\label{app:secondary}

\rev{The two ablations in this section were run with a single seed and are reported as secondary observations only; the seeded comparisons of Appendix~\ref{app:seeds} do not depend on them.}

\paragraph{Context size.}
We evaluated every curriculum with \ticl{} context sizes 512, 1024, and 2048. For every curriculum, the 1024 configuration outperformed both its 512 and its 2048 counterpart, so we report 1024 throughout. We read this as a sweet spot between the informativeness of the in-context set and the effective context utilization of the frozen head: 512 examples appear to under-specify the tasks, while 2048 did not convert the additional examples into better predictions.

\paragraph{Early-stopping patience.}
The reported $\mathrm{PT}_{\sg}$ run uses a patience of 40 evaluations. An otherwise identical run with a patience of 20 evaluations performed worse across the board. The explanation is mechanical: with a longer patience the run survives longer plateaus and therefore has more opportunities to save a better best-on-validation checkpoint. Since checkpoint selection is best-on-validation, extra patience can only match or improve the selected checkpoint at the cost of compute. We consequently used patience values of 20 to 30 evaluations for the longer multi-stage runs as a compute compromise.

\section[Data Construction and Reproducibility]{\rev{Data Construction and Reproducibility}}
\label{app:data}

\begin{table}[!htb]
\centering
\caption{Synthetic pretraining corpora. \sa{} $=$ B $\cup$ C $\cup$ D (9{,}143 DBs); \sg{} $=$ SALT (500 DBs). \rev{(Moved from Section~\ref{sec:setup}; content unchanged.)}}
\label{tab:corpora}
\small
\begin{tabular}{lcccc}
\toprule
 & \textbf{B} & \textbf{C} & \textbf{D} & \textbf{SALT (\sg{})} \\
\midrule
Databases & 5{,}623 & 1{,}408 & 2{,}112 & 500 \\
Tables per DB (min/max/avg) & 2\,/\,4\,/\,3.0 & 2\,/\,4\,/\,3.0 & 5\,/\,8\,/\,6.5 & 4\,/\,4\,/\,4 \\
Total rows per DB (min--max) & 150--600 & 50--400 & 50--400 & 500--5{,}000 \\
Columns per table (min/max/avg) & 2\,/\,13\,/\,7 & 4\,/\,20\,/\,10 & 4\,/\,22\,/\,7 & 2\,/\,13\,/\,7 \\
Tasks per DB & $\sim$6 (binary) & $\sim$6 (binary) & $\sim$6 (binary) & 7 (multi-class) \\
\bottomrule
\end{tabular}
\end{table}

\begin{revblock}
\paragraph{Schema-guided corpus (\sg{}).}
The 500 \sg{} databases share the schema of the real \texttt{rel-salt} SAP sales database: its four tables and its real FK graph. Table content is generated by PluRel~\citep{abolhasani2026plurel} and then post-processed so that table cardinalities match the real database (500 to 5{,}000 rows per database, Table~\ref{tab:corpora}). The 7 tasks per database predict genuine categorical SAP fields of the sales documents and items, and the rows of every database are split temporally, with training rows dated before 1~February~2020, validation rows before 1~July~2020, and test rows afterwards.

\paragraph{Schema-agnostic corpus (\sa{}).}
The 9{,}143 \sa{} databases come from one PluRel generator that draws a fresh schema for every database; sub-families B, C, and D are the same generator run under three configuration settings that control the number of tables, the FK depth, and the row and column budgets (Table~\ref{tab:corpora}), not three separate generators. Every task instance masks one feature column of an entity table and predicts it from the row's remaining columns and its FK neighborhood; targets are binarized ($k{=}2$). Rows are split deterministically into 75\,\% training, 10\,\% validation, and 15\,\% test.

\paragraph{Regenerability.}
Each synthetic database is a deterministic function of an integer seed and a fixed generator configuration, so both corpora, and every subset used in this paper (including the 1{,}400-database \sa{} subset of the unordered controls, which is selected by a pool shuffle with seed 42), can be regenerated exactly from the generator scripts without storing the data.
\end{revblock}

\section{Additional Execution Details}
\label{app:execution}

\paragraph{Training and checkpoint-selection protocol. \rev{(Moved from Section~\ref{sec:setup}; content unchanged.)}}
Every pretraining stage uses the same best-on-validation protocol. We hold out a stratified validation split (fraction $10^{-3}$ across all tasks, fixed split seed), evaluate every 250 steps, use threshold-free AUROC for model selection on binary tasks, and apply early stopping with a patience of 20--30 evaluations after a 5{,}000-step floor. The best validation checkpoint is carried to the next stage or downstream evaluation. Synthetic pretraining binarizes targets ($k{=}2$), whereas real-data pretraining retains the original classes ($k{=}0$). Table~\ref{tab:config} lists the shared configuration. All experiments ran on Azure \texttt{Standard\_NC80adis\_H100\_v5} instances (2$\times$H100, 80 vCPUs, 640\,GB RAM) with 2-GPU distributed training.

\begin{table}[!htb]
\centering
\caption{Shared configuration across all pretraining stages and frozen \ticl{} evaluation. \rev{(Moved from Section~\ref{sec:setup}; content unchanged.)}}
\label{tab:config}
\small
\begin{tabular}{llll}
\toprule
\multicolumn{2}{c}{\textbf{Encoder / sampling}} & \multicolumn{2}{c}{\textbf{Optimization / evaluation}} \\
\midrule
Message-passing layers & 4 & Batch size & 256 \\
Hidden dim.\ & 512 & Learning rate / weight decay & $3\!\times\!10^{-4}$ / $2\!\times\!10^{-4}$ \\
Subgraph hops & 2 & Max / min steps & 100k / 5k \\
Fanout / few-shot fanout & 20 / 3 & Eval.\ interval / patience & 250 / 20--30 \\
Reversed FK edges & yes & Binarization (synth.\ / real) & $k{=}2$ / $k{=}0$ \\
Gating & no & \ticl{} (v2) ensemble / context & 8 / $\{512, 1024, 2048\}$ \\
\bottomrule
\end{tabular}
\end{table}

\paragraph{Multi-database streaming.}
Pretraining iterates over thousands of preprocessed databases. Because the full corpus does not fit in memory, at most 1{,}000 databases are resident at a time under a least-recently-used eviction policy. This constraint is particularly relevant for the schema-guided SALT corpus, whose item-level tasks generate substantially more training subgraphs per database than B, C, or D.

\paragraph{Validation and stage chaining.}
Validation is constructed from a fraction $10^{-3}$ of training rows using a fixed split seed (12{,}345), stratified across tasks. Validation AUROC for binary targets is evaluated every 250 steps for checkpoint selection. Early stopping occurs after 20--30 evaluations without improvement, subject to a 5{,}000-step minimum. Each curriculum stage initializes from the best validation checkpoint of the preceding stage. Data order is fixed using a shuffle seed of 42. Synthetic pretraining uses binarized targets ($k=2$), whereas real-data pretraining retains the original target cardinality. All stages use the shared encoder and optimization configuration reported in Table~\ref{tab:config}. \rev{A stage is launched as an independent pretraining job that loads the previous stage's best checkpoint and otherwise uses the identical step ceiling (100k), floor (5k), evaluation interval, patience, batch size, and learning rate, so a multi-stage curriculum differs from a single-stage run only in the data each stage sees. The unordered controls are single stages over the shuffled pool of all 500 \sg{} databases and a fixed subset of 1{,}400 \sa{} databases (pool shuffle seed 42), plus the corresponding real families where applicable.}

\paragraph{\rev{Seeds.}}
\rev{Every configuration is run with five seeds. The seed governs the model initialization and the mini-batch order of every stage of the curriculum, the random weights of the untrained \textsc{RandGriffin} baseline, and the sampling of the in-context set during evaluation; the database-pool shuffle seed (42) and the validation split seed (12{,}345) are held fixed, so that all seeds see the same data. Reported values are the mean and standard deviation of the task-averaged ROC-AUC over the five end-to-end runs.}

\paragraph{Frozen downstream evaluation.}
After pretraining, the selected Griffin checkpoint is frozen and used to generate relational representations for each downstream task. These representations are passed to a frozen TabICL-v2 predictor with an ensemble of eight estimators. We evaluate context sizes of 512, 1{,}024, and 2{,}048; the main results use 1{,}024, which performs best across the evaluated curricula. Neither Griffin nor TabICL undergoes gradient-based adaptation on the downstream tasks. \rev{Concretely, for each task the frozen encoder embeds every target row from a 2-hop temporal subgraph sampled with the pretraining settings (fanout 20, few-shot fanout 3, four message-passing layers, hidden size 512, reversed FK edges). The in-context set consists of $c$ labeled rows sampled from the task's training split ($c{=}1024$ in the main results), the query set is the task's held-out rows, and \ticl{} scores the queries with its eight-estimator ensemble on a single GPU. The two baselines run the identical evaluation script: \textsc{RandGriffin} skips loading a checkpoint and embeds with randomly initialized weights, and \textsc{RawFeat} bypasses the encoder and its projection into the head's feature space and feeds the target row's own features to \ticl{}.}

\end{document}